%% file: deepqnet.tex
\documentclass{article} 
\usepackage{nips13submit_e,times}
\usepackage{hyperref}
\usepackage{url}
\usepackage{amsfonts}
\usepackage{amsmath}
\usepackage{algorithm}
\usepackage{algorithmicx}
\usepackage{algpseudocode}
\usepackage{graphicx}

\title{Playing Atari with Deep Reinforcement Learning}

\author{
Volodymyr Mnih \hspace{0.3cm}
Koray Kavukcuoglu \hspace{0.3cm}
David Silver \hspace{0.3cm}
Alex Graves \hspace{0.3cm} 
Ioannis Antonoglou\\
\\
{\bf Daan Wierstra \hspace{0.3cm}
Martin Riedmiller}\\
\\
DeepMind Technologies\\
\\
\small{\texttt{ \{vlad,koray,david,alex.graves,ioannis,daan,martin.riedmiller\} @ deepmind.com }}
}

\nipsfinalcopy 

\begin{document}
\maketitle

\begin{abstract}
\input{abstract}
\end{abstract}

\newcommand{\ssq}{s^{(t)}}
\newcommand{\asq}{a^{(t)}}
\newcommand{\pasq}{a^{(t-1)}}
\newcommand{\stpn}{s_{t+1}}
\newcommand{\qpi}{Q^{\pi}\left(\ssq,a_t\right)}
\newcommand{\vpi}{V^{\pi}\left(\ssq\right)}
\newcommand{\aqpi}{\hat{Q}^{\pi}}

\newcommand{\prob}[1]{\mathbb{P} \left[ #1 \right]}
\newcommand{\given}{|}
\newcommand{\expect}[1]{\mathbb{E} \left[ #1 \right]}
\newcommand{\expectx}[2]{\mathbb{E}_{#1} \left[ #2 \right]}
\newcommand{\eqdef}{\stackrel{def}{=}}

\section{Introduction}
\label{sec:introduction}

\input{intro}

\section{Background}
\label{sec:background}
\input{background}

\section{Deep Reinforcement Learning}
\label{sec:method}
\input{method}

\section{Experiments}
\label{sec:experiments}
\input{experiments}

\vspace{-0.3cm}
\section{Conclusion}
\vspace{-0.2cm}
\label{sec:conclusion}
\input{conclusion}
\bibliographystyle{plain}
\bibliography{deepqnet}

\end{document}

%% file: abstract.tex

We present the first deep learning model to successfully learn control policies directly from high-dimensional sensory input using reinforcement learning. 
The model is a convolutional neural network, trained with a variant of Q-learning, whose input is raw pixels and whose output is a value function estimating future rewards.
We apply our method to seven Atari 2600 games from the Arcade Learning Environment, with no adjustment of the architecture or learning algorithm.
We find that it outperforms all previous approaches on six of the games and surpasses a human expert on three of them.
%
%

%% file: intro.tex
Learning to control agents directly from high-dimensional sensory inputs like vision and speech is one of the long-standing challenges of reinforcement learning (RL). 
Most successful RL applications that operate on these domains have relied on hand-crafted features combined with linear value functions or policy representations. 
Clearly, the performance of such systems heavily relies on the quality of the feature representation. 

Recent advances in deep learning have made it possible to extract high-level features from raw sensory data, leading to breakthroughs in computer vision~\cite{krizhevsky-imagenet, sermanet-cvpr-2013, mnih-thesis} and speech recognition~\cite{dahl-speech,graves-speech}.
These methods utilise a range of neural network architectures, including convolutional networks, multilayer perceptrons, restricted Boltzmann machines and recurrent neural networks, and have exploited both supervised and unsupervised learning. 
It seems natural to ask whether similar techniques could also be beneficial for RL with sensory data.

%
%
%

However reinforcement learning presents several challenges from a deep learning perspective.
Firstly, most successful deep learning applications to date have required large amounts of hand-labelled training data.
RL algorithms, on the other hand, must be able to learn from a scalar reward signal that is frequently sparse, noisy and delayed.
The delay between actions and resulting rewards, which can be thousands of timesteps long, seems particularly daunting when compared to the direct association between inputs and targets found in supervised learning.
Another issue is that most deep learning algorithms assume the data samples to be independent, while in reinforcement learning one typically encounters sequences of highly correlated states.
Furthermore, in RL the data distribution changes as the algorithm learns new behaviours, which can be problematic for deep learning methods that assume a fixed underlying distribution.


This paper demonstrates that a convolutional neural network can overcome these challenges to learn successful control policies from raw video data in complex RL environments.
The network is trained with a variant of the Q-learning~\cite{watkins-qlearning} algorithm, with stochastic gradient descent to update the weights.
To alleviate the problems of correlated data and non-stationary distributions, we use an experience replay mechanism~\cite{lin1993reinforcement} which randomly samples previous transitions, and thereby smooths the training distribution over many past behaviors.

\begin{figure}
\includegraphics[width=0.195\linewidth]{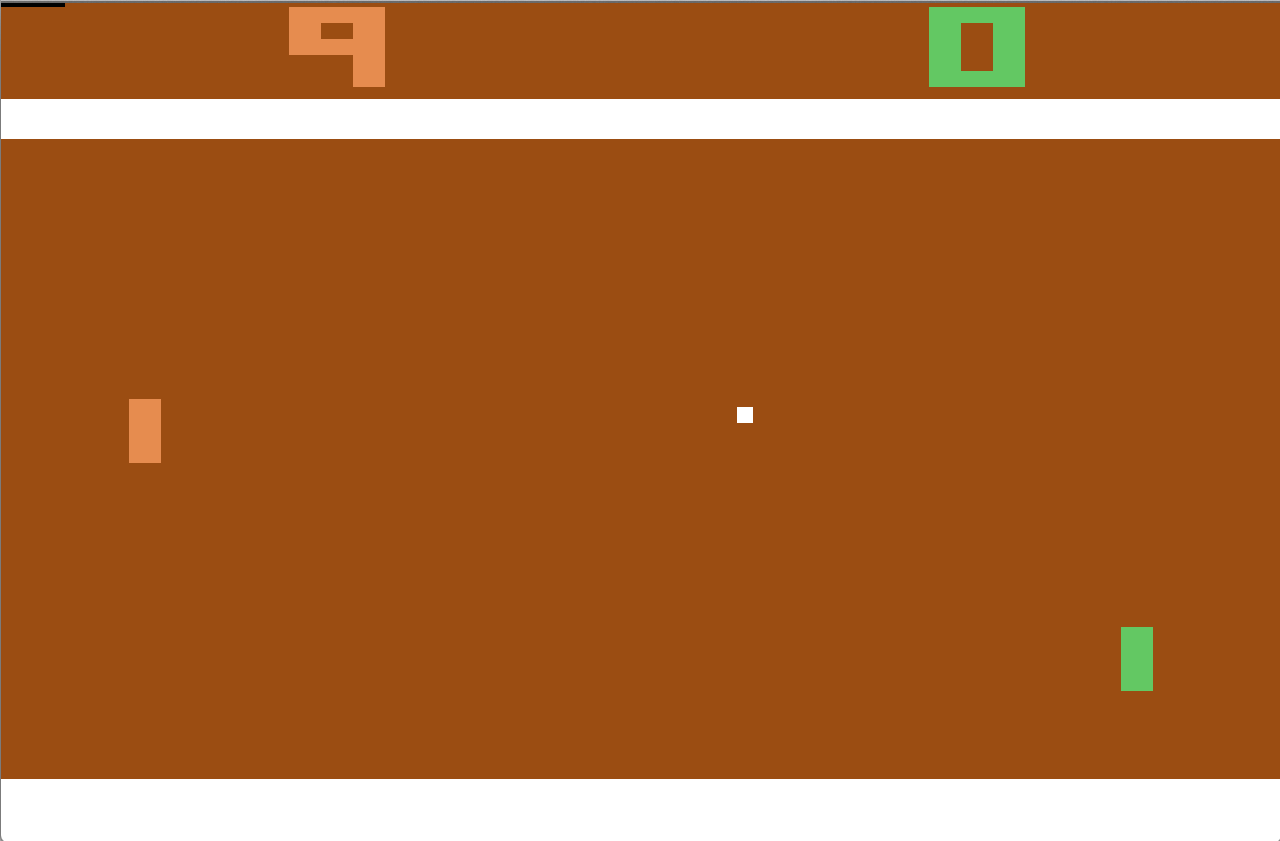}
\includegraphics[width=0.195\linewidth]{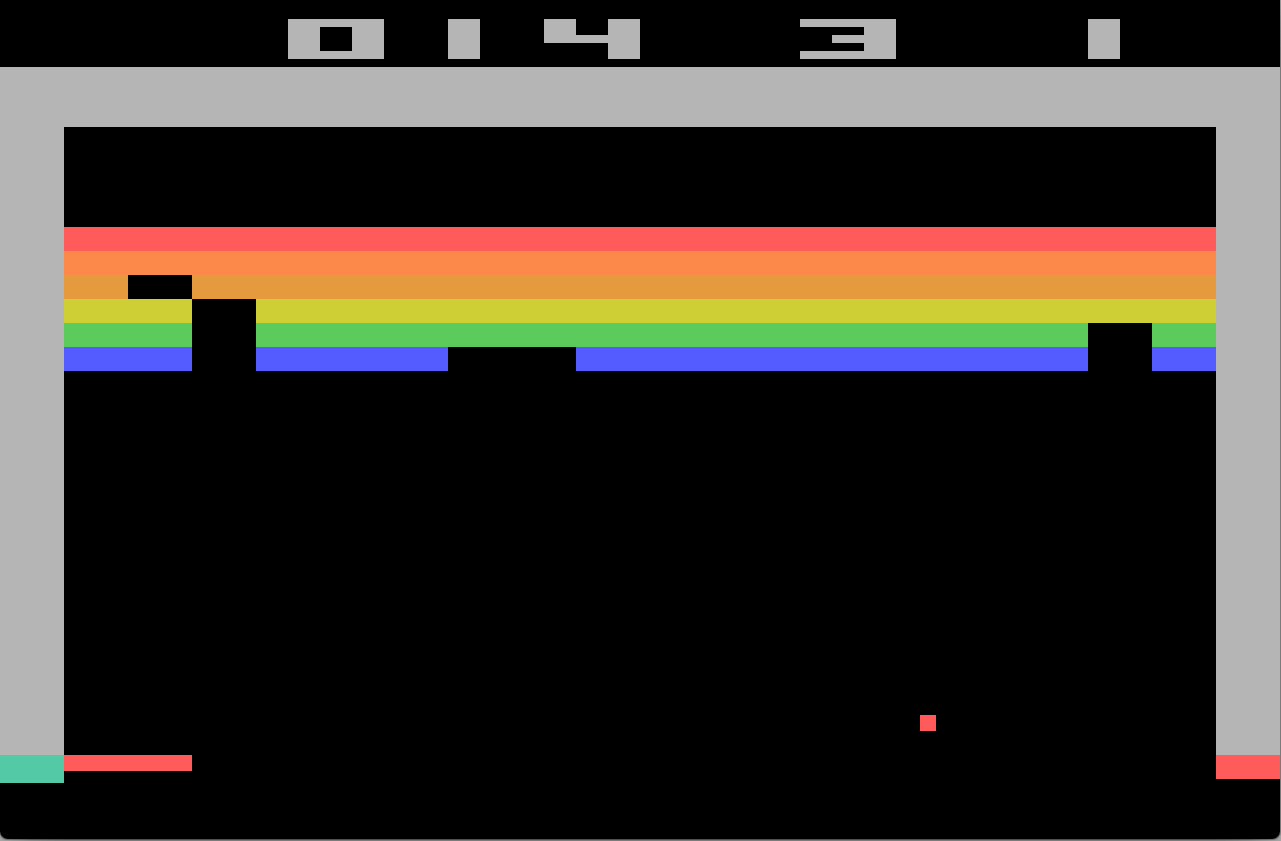}
\includegraphics[width=0.195\linewidth]{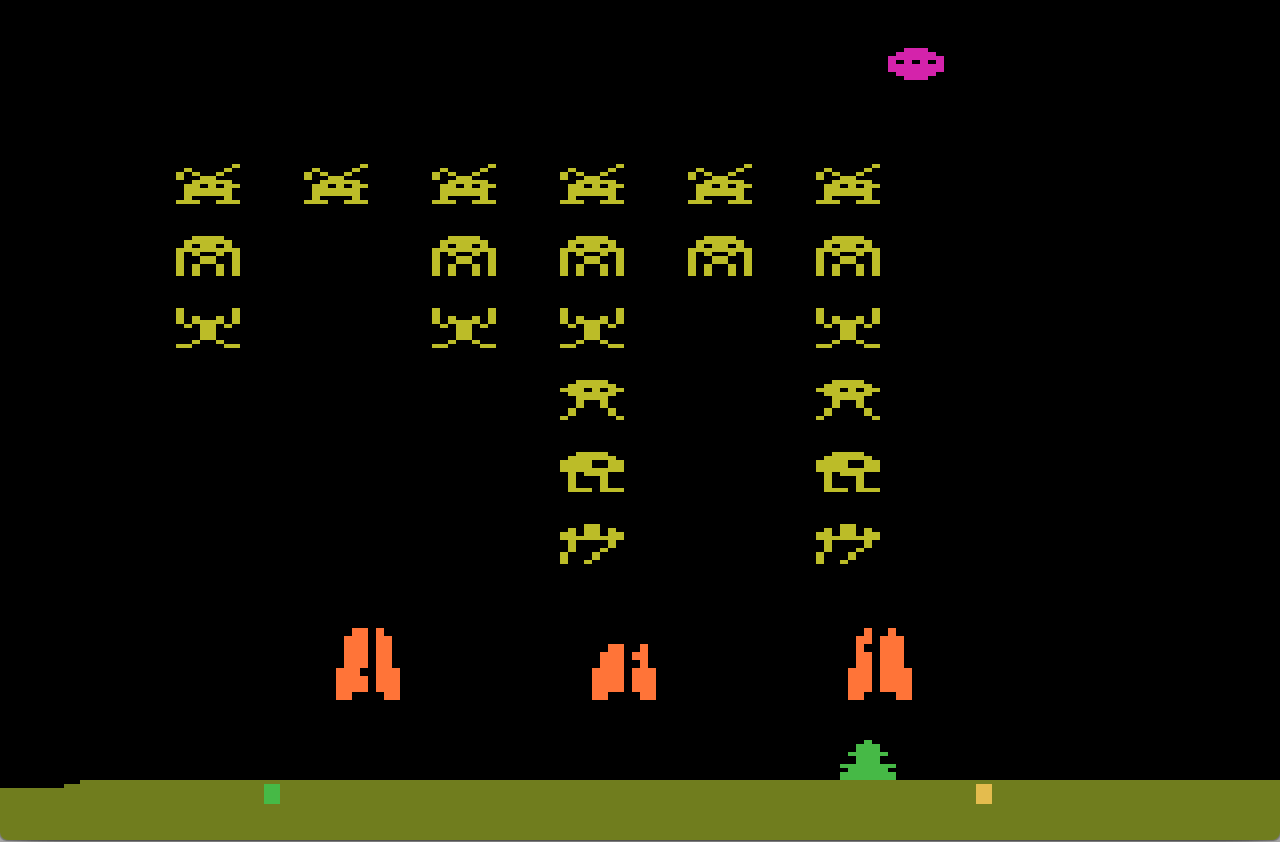}
\includegraphics[width=0.195\linewidth]{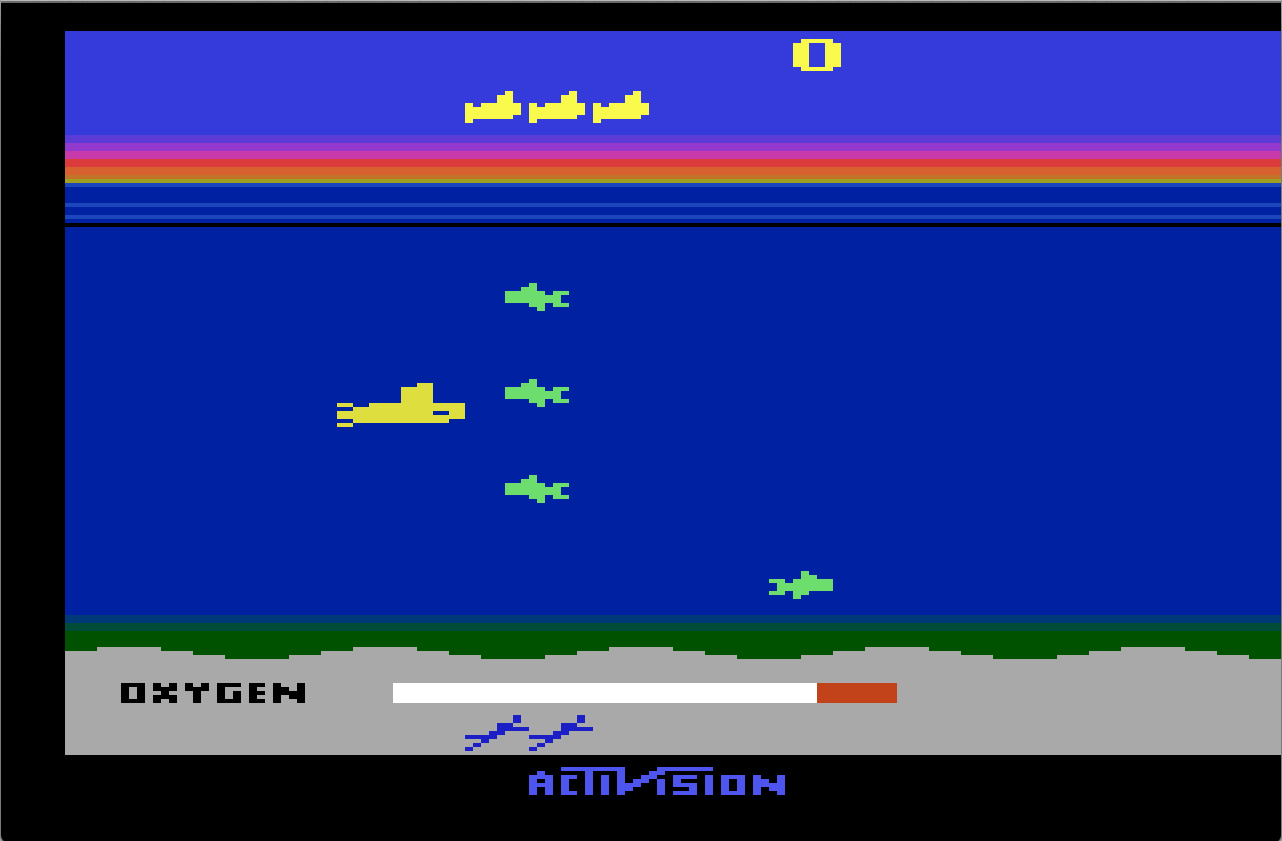}
\includegraphics[width=0.195\linewidth]{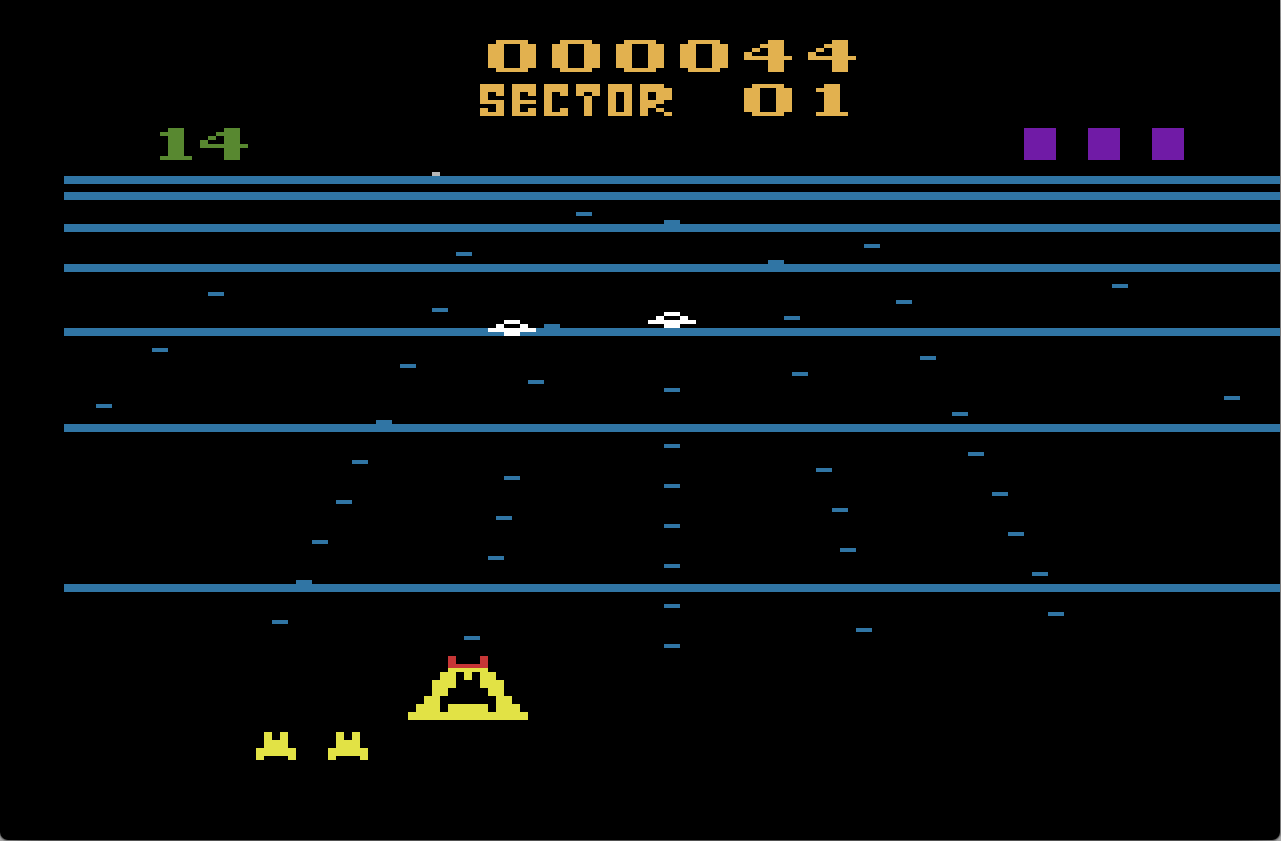}
\caption{\label{fig-games} Screen shots from five Atari 2600 Games: (\emph{Left-to-right}) Pong, Breakout, Space Invaders, Seaquest, Beam Rider}
\end{figure}

We apply our approach to a range of Atari 2600 games implemented in The Arcade Learning Environment (ALE)~\cite{bellemare-ale}.
Atari 2600 is a challenging RL testbed that presents agents with a high dimensional visual input ($210 \times 160$ RGB video at 60Hz) and a diverse and interesting set of tasks that were designed to be difficult for humans players.
%
%
Our goal is to create a single neural network agent that is able to successfully learn to play as many of the games as possible.
The network was not provided with any game-specific information or hand-designed visual features, and was not privy to the internal state of the emulator; it learned from nothing but the video input, the reward and terminal signals, and the set of possible actions---just as a human player would.
Furthermore the network architecture and all hyperparameters used for training were kept constant across the games.
So far the network has outperformed all previous RL algorithms on six of the seven games we have attempted and surpassed an expert human player on three of them.
Figure~\ref{fig-games} provides sample screenshots from five of the games used for training.

%% file: background.tex

We consider tasks in which an agent interacts with an environment $\mathcal{E}$, in this case the Atari emulator, in a sequence of actions, observations and rewards. 
At each time-step the agent selects an action $a_t$ from the set of legal game actions, $\mathcal{A}=\{1, \ldots, K \}$. The action is passed to the emulator and modifies its internal state and the game score. In general $\mathcal{E}$ may be stochastic. The emulator's internal state is not observed by the agent; instead it observes an image $x_t \in \mathbb{R}^d$ from the emulator, which is a vector of raw pixel values representing the current screen. In addition it receives a reward $r_t$ representing the change in game score. Note that in general the game score may depend on the whole prior sequence of actions and observations; feedback about an action may only be received after many thousands of time-steps have elapsed. 

Since the agent only observes images of the current screen, the task is partially observed and many emulator states are perceptually aliased, i.e. it is impossible to fully understand the current situation from only the current screen $x_t$. We therefore consider sequences of actions and observations, $s_t = {x_1, a_1, x_2, ..., a_{t-1}, x_t}$, and learn game strategies that depend upon these sequences. All sequences in the emulator are assumed to terminate in a finite number of time-steps. This formalism gives rise to a large but finite Markov decision process (MDP) in which each sequence is a distinct state. As a result, we can apply standard reinforcement learning methods for MDPs, simply by using the complete sequence $s_t$ as the state representation at time $t$.

The goal of the agent is to interact with the emulator by selecting actions in a way that maximises future rewards. We make the standard assumption that future rewards are discounted by a factor of $\gamma$ per time-step, and define the future discounted \emph{return} at time $t$ as $R_t = \sum_{t'=t}^{T} \gamma^{t'-t} r_{t'}$, where $T$
is the time-step at which the game terminates. We define the optimal action-value function $Q^*(s,a)$ as the maximum expected return achievable by following any strategy, after seeing some sequence $s$ and then taking some action $a$, $Q^*(s,a) = \max_{\pi} \expect{ R_t | s_t=s, a_t=a, \pi }$, where $\pi$ is a policy mapping sequences to actions (or distributions over actions).

The optimal action-value function obeys an important identity known as the \emph{Bellman equation}. This is based on the following intuition: if the optimal value $Q^*(s',a')$ of the sequence $s'$ at the next time-step was known for all possible actions $a'$, then the optimal strategy is to select the action $a'$ maximising the expected value of $r + \gamma Q^*(s',a')$,
\begin{align}
Q^*(s,a) &= \expectx{s' \sim \mathcal{E}}{ r + \gamma \max_{a'} Q^*(s', a') \Big| s, a }
\end{align}
The basic idea behind many reinforcement learning algorithms is to estimate the action-value function, by using the Bellman equation as an iterative update, $Q_{i+1}(s,a) = \mathbb{E}\left[ r + \gamma \max_{a'} Q_i(s', a') | s, a \right]$. Such \emph{value iteration} algorithms converge to the optimal action-value function, $Q_i \rightarrow Q^*$ as $i \rightarrow \infty$ \cite{sutton:book}. In practice, this basic approach is totally impractical, because the action-value function is estimated separately for each sequence, without any generalisation. Instead, it is common to use a function approximator to estimate the action-value function, $Q(s,a; \theta) \approx Q^*(s,a)$. In the reinforcement learning community this is typically a linear function approximator, but sometimes a non-linear function approximator is used instead, such as a neural network. We refer to a neural network function approximator with weights $\theta$ as a Q-network. A Q-network can be trained by minimising a sequence of loss functions $L_i(\theta_i)$ that changes at each iteration $i$,
\begin{align}
L_i\left(\theta_i\right) &= \expectx{s,a \sim \rho(\cdot)}{\left(y_i - Q \left(s,a ; \theta_i \right) \right)^2},
\label{eq:q-learning-loss}
\end{align}
where $y_i = \expectx{s' \sim \mathcal{E}}{r + \gamma \max_{a'} Q(s', a'; \theta_{i-1}) | s, a }$ is the target for iteration $i$ and $\rho(s,a)$ is a probability distribution over sequences $s$ and actions $a$ that we refer to as the \emph{behaviour distribution}. The parameters from the previous iteration $\theta_{i-1}$ are held fixed when optimising the loss function $L_i\left(\theta_i\right)$. 
Note that the targets depend on the network weights; this is in contrast with the targets used for supervised learning, which are fixed before learning begins.
Differentiating the loss function with respect to the weights we arrive at the following gradient,
\begin{align}
\nabla_{\theta_i} L_i\left(\theta_i\right) &= \mathbb{E}_{s,a \sim \rho(\cdot); s' \sim \mathcal{E}} \left[ \left( r + \gamma \max_{a'} Q(s', a'; \theta_{i-1}) - Q(s,a ; \theta_i ) \right) \nabla_{\theta_i} Q(s,a;\theta_{i}) \right] .
\label{eq:q-learning-gradient}
\end{align}
Rather than computing the full expectations in the above gradient, it is often computationally expedient to optimise the loss function by stochastic gradient descent. If the weights are updated after every time-step, and the expectations are replaced by single samples from the behaviour distribution $\rho$ and the emulator $\mathcal{E}$ respectively, then we arrive at the familiar \emph{Q-learning} algorithm \cite{watkins-qlearning}. 

Note that this algorithm is \emph{model-free}: it solves the reinforcement learning task directly using samples from the emulator $\mathcal{E}$, without explicitly constructing an estimate of $\mathcal{E}$. It is also \emph{off-policy}: it learns about the greedy strategy $a = \max_{a} Q(s,a;\theta)$, while following a behaviour distribution that ensures adequate exploration of the state space. In practice, the behaviour distribution is often selected by an $\epsilon$-greedy strategy that follows the greedy strategy with probability $1 - \epsilon$ and selects a random action with probability $\epsilon$.

\section{Related Work}

Perhaps the best-known success story of reinforcement learning is \emph{TD-gammon}, a backgammon-playing program which learnt entirely by reinforcement learning and self-play, and achieved a super-human level of play~\cite{tesauro-95-backgammon}. TD-gammon used a model-free reinforcement learning algorithm similar to Q-learning, and approximated the value function using a multi-layer perceptron with one hidden layer\footnote{In fact TD-Gammon approximated the state value function $V(s)$ rather than the action-value function $Q(s,a)$, and learnt \emph{on-policy} directly from the self-play games}. 

However, early attempts to follow up on TD-gammon, including applications of the same method to chess, Go and checkers were less successful. This led to a widespread belief that the TD-gammon approach was a special case that only worked in backgammon, perhaps because the stochasticity in the dice rolls helps explore the state space and also makes the value function particularly smooth \cite{pollack:td-gammon}.

Furthermore, it was shown that combining model-free reinforcement learning algorithms such as Q-learning with non-linear function approximators \cite{tsitsiklis:td-convergence}, or indeed with off-policy learning \cite{baird:residual} could cause the Q-network to diverge. Subsequently, the majority of work in reinforcement learning focused on linear function approximators with better convergence guarantees \cite{tsitsiklis:td-convergence}.

More recently, there has been a revival of interest in combining deep learning with reinforcement learning. Deep neural networks have been used to estimate the environment $\mathcal{E}$; restricted Boltzmann machines have been used to estimate the value function \cite{sallans:rbm}; or the policy \cite{heess:energy-based}. In addition, the divergence issues with Q-learning have been partially addressed by \emph{gradient temporal-difference} methods. These methods are proven to converge when evaluating a fixed policy with a nonlinear function approximator \cite{maei:nonlinear}; or when learning a control policy with linear function approximation using a restricted variant of Q-learning \cite{maei:gq}. However, these methods have not yet been extended to nonlinear control.

Perhaps the most similar prior work to our own approach is neural fitted Q-learning (NFQ) \cite{riedmiller-nfq}. NFQ optimises the sequence of loss functions in Equation \ref{eq:q-learning-loss}, using the RPROP algorithm to update the parameters of the Q-network. However, it uses a batch update that has a computational cost per iteration that is proportional to the size of the data set, whereas we consider stochastic gradient updates that have a low constant cost per iteration and scale to large data-sets. NFQ has also been successfully applied to simple real-world control tasks using purely visual input, by first using deep autoencoders to learn a low dimensional representation of the task, and then applying NFQ to this representation \cite{lange:dfq}. In contrast our approach applies reinforcement learning end-to-end, directly from the visual inputs; as a result it may learn features that are directly relevant to discriminating action-values.  Q-learning has also previously been combined with experience replay and a simple neural network \cite{lin1993reinforcement}, but again starting with a low-dimensional state rather than raw visual inputs.

The use of the Atari 2600 emulator as a reinforcement learning platform was introduced by \cite{bellemare-ale}, who applied standard reinforcement learning algorithms with linear function approximation and generic visual features. Subsequently, results were improved by using a larger number of features, and using tug-of-war hashing to randomly project the features into a lower-dimensional space~\cite{bellemare2012sketch}. The HyperNEAT evolutionary architecture \cite{hausknecht-neuro} has also been applied to the Atari platform, where it was used to evolve (separately, for each distinct game) a neural network representing a strategy for that game. When trained repeatedly against deterministic sequences using the emulator's reset facility, these strategies were able to exploit design flaws in several Atari games.

%% file: method.tex

Recent breakthroughs in computer vision and speech recognition have relied on efficiently training deep neural networks on very large training sets.
The most successful approaches are trained directly from the raw inputs, using lightweight updates based on stochastic gradient descent. By feeding sufficient data into deep neural networks, it is often possible to learn better representations than handcrafted features \cite{krizhevsky-imagenet}. These successes motivate our approach to reinforcement learning. Our goal is to connect a reinforcement learning algorithm to a deep neural network which operates directly on RGB images and efficiently process training data by using stochastic gradient updates. 

Tesauro's TD-Gammon architecture provides a starting point for such an approach. This architecture updates the parameters of a network that estimates the value function, directly from on-policy samples of experience, $s_t, a_t, r_t, s_{t+1}, a_{t+1}$, drawn from the algorithm's interactions with the environment (or by self-play, in the case of backgammon). Since this approach was able to outperform the best human backgammon players 20 years ago, it is natural to wonder whether two decades of hardware improvements, coupled with modern deep neural network architectures and scalable RL algorithms might produce significant progress. 

In contrast to TD-Gammon and similar online approaches, we utilize a technique known as \emph{experience replay}~\cite{lin1993reinforcement} where we store the agent's experiences at each time-step, $e_t = (s_t, a_t, r_t, s_{t+1})$ in a data-set $\mathcal{D} = e_1, ..., e_N$, pooled over many episodes into a \emph{replay memory}. During the inner loop of the algorithm, we apply Q-learning updates, or minibatch updates, to samples of experience, $e \sim \mathcal{D}$, drawn at random from the pool of stored samples. After performing experience replay, the agent selects and executes an action according to an $\epsilon$-greedy policy. Since using histories of arbitrary length as inputs to a neural network can be difficult, our Q-function instead works on fixed length representation of histories produced by a function $\phi$. The full algorithm, which we call \emph{deep Q-learning}, is presented in Algorithm~\ref{alg}.

This approach has several advantages over standard online Q-learning~\cite{sutton:book}. First, each step of experience is potentially used in many weight updates, which allows for greater data efficiency. Second, learning directly from consecutive samples is inefficient, due to the strong correlations between the samples; randomizing the samples breaks these correlations and therefore reduces the variance of the updates. Third, when learning on-policy the current parameters determine the next data sample that the parameters are trained on. For example, if the maximizing action is to move left then the training samples will be dominated by samples from the left-hand side; if the maximizing action then switches to the right then the training distribution will also switch. It is easy to see how unwanted feedback loops may arise and the parameters could get stuck in a poor local minimum, or even diverge catastrophically \cite{tsitsiklis:td-convergence}. By using experience replay the behavior distribution is averaged over many of its previous states, smoothing out learning and avoiding oscillations or divergence in the parameters. Note that when learning by experience replay, it is necessary to learn off-policy (because our current parameters are different to those used to generate the sample), which motivates the choice of Q-learning.

In practice, our algorithm only stores the last $N$ experience tuples in the replay memory, and samples uniformly at random from $\mathcal{D}$ when performing updates. This approach is in some respects limited since the memory buffer does not differentiate important transitions and always overwrites with recent transitions due to the finite memory size $N$. Similarly, the uniform sampling gives equal importance to all transitions in the replay memory. A more sophisticated sampling strategy might emphasize transitions from which we can learn the most, similar to prioritized sweeping \cite{moore:prioritized}.  %


\begin{algorithm}[t]
\begin{algorithmic}
\State Initialize replay memory $\mathcal{D}$ to capacity $N$
\State Initialize action-value function $Q$ with random weights
\For{episode $=1,M$} 
\State Initialise sequence $s_1 = \{x_1\}$ and preprocessed sequenced $\phi_1 = \phi(s_1)$
\For {$t=1,T$}
	\State With probability $\epsilon$ select a random action $a_t$
	\State otherwise select $a_t = \max_{a} Q^*(\phi(s_t), a; \theta)$
	\State Execute action $a_t$ in emulator and observe reward $r_t$ and image $x_{t+1}$
	\State Set $s_{t+1} = s_t,a_t,x_{t+1}$ and preprocess $\phi_{t+1} = \phi(s_{t+1})$
	\State Store transition $\left(\phi_t,a_t,r_t,\phi_{t+1}\right)$ in $\mathcal{D}$
	\State Sample random minibatch of transitions $\left(\phi_j,a_j,r_j,\phi_{j+1}\right)$ from $\mathcal{D}$
	\State Set
	$y_j =
    \left\{
    \begin{array}{l l}
      r_j  \quad & \text{for terminal } \phi_{j+1}\\
      r_j + \gamma \max_{a'} Q(\phi_{j+1}, a'; \theta) \quad & \text{for non-terminal } \phi_{j+1}
    \end{array} \right.$
	\State Perform a gradient descent step on $\left(y_j - Q(\phi_j, a_j; \theta) \right)^2$ according to equation~\ref{eq:q-learning-gradient}
\EndFor
\EndFor
\end{algorithmic}
\caption{Deep Q-learning with Experience Replay}
\label{alg}
\end{algorithm}

\subsection{Preprocessing and Model Architecture}

Working directly with raw Atari frames, which are $210\times 160$ pixel images with a 128 color palette, can be computationally demanding, so we apply a basic preprocessing step aimed at reducing the input dimensionality.  The raw frames are preprocessed by first converting their RGB representation to gray-scale and down-sampling it to a $110 \times 84$ image. The final input representation is obtained by cropping an $84\times 84$ region of the image that roughly captures the playing area. The final cropping stage is only required because we use the GPU implementation of 2D convolutions from~\cite{krizhevsky-imagenet}, which expects square inputs.  For the experiments in this paper, the function $\phi$ from algorithm~\ref{alg} applies this preprocessing to the last $4$ frames of a history and stacks them to produce the input to the $Q$-function. 

There are several possible ways of parameterizing $Q$ using a neural network.  Since $Q$ maps history-action pairs to scalar estimates of their Q-value, the history and the action have been used as inputs to the neural network by some previous approaches~\cite{riedmiller-nfq,lange:dfq}.  The main drawback of this type of architecture is that a separate forward pass is required to compute the Q-value of each action, resulting in a cost that scales linearly with the number of actions.  We instead use an architecture in which there is a separate output unit for each possible action, and only the state representation is an input to the neural network.  The outputs correspond to the predicted Q-values of the individual action for the input state.  The main advantage of this type of architecture is the ability to compute Q-values for all possible actions in a given state with only a single forward pass through the network. 

We now describe the exact architecture used for all seven Atari games.
The input to the neural network consists is an $84 \times 84 \times 4$ image produced by $\phi$.  The first hidden layer convolves $16$ $8 \times 8$ filters with stride $4$ with the input image and applies a rectifier nonlinearity~\cite{jarrett-best,nair-relu}.  The second hidden layer convolves $32$ $4\times 4$ filters with stride $2$, again followed by a rectifier nonlinearity.  The final hidden layer is fully-connected and consists of $256$ rectifier units.  The output layer is a fully-connected linear layer with a single output for each valid action.  The number of valid actions varied between $4$ and $18$ on the games we considered.  We refer to convolutional networks trained with our approach as Deep Q-Networks (DQN).

%% file: experiments.tex

So far, we have performed experiments on seven popular ATARI games -- Beam Rider, Breakout, Enduro, Pong, Q*bert, Seaquest, Space Invaders. We use the same network architecture, learning algorithm and hyperparameters settings across all seven games, showing that our approach is robust enough to work on a variety of games without incorporating game-specific information. While we evaluated our agents on the real and unmodified games, we made one change to the reward structure of the games during training only.  Since the scale of scores varies greatly from game to game, we fixed all positive rewards to be $1$ and all negative rewards to be $-1$, leaving $0$ rewards unchanged.  Clipping the rewards in this manner limits the scale of the error derivatives and makes it easier to use the same learning rate across multiple games.  At the same time, it could affect the performance of our agent since it cannot differentiate between rewards of different magnitude.

In these experiments, we used the RMSProp algorithm with minibatches of size 32.  The behavior policy during training was $\epsilon$-greedy with $\epsilon$ annealed linearly from $1$ to $0.1$ over the first million frames, and fixed at $0.1$ thereafter.  We trained for a total of $10$ million frames and used a replay memory of one million most recent frames.

Following previous approaches to playing Atari games, we also use a simple frame-skipping technique~\cite{bellemare-ale}. More precisely, the agent sees and selects actions on every $k^{th}$ frame instead of every frame, and its last action is repeated on skipped frames.  Since running the emulator forward for one step requires much less computation than having the agent select an action, this technique allows the agent to play roughly $k$ times more games without significantly increasing the runtime.  We use $k=4$ for all games except Space Invaders where we noticed that using $k=4$ makes the lasers invisible because of the period at which they blink.  We used $k=3$ to make the lasers visible and this change was the only difference in hyperparameter values between any of the games.

\subsection{Training and Stability}

\begin{figure}
\includegraphics[width=0.245\linewidth]{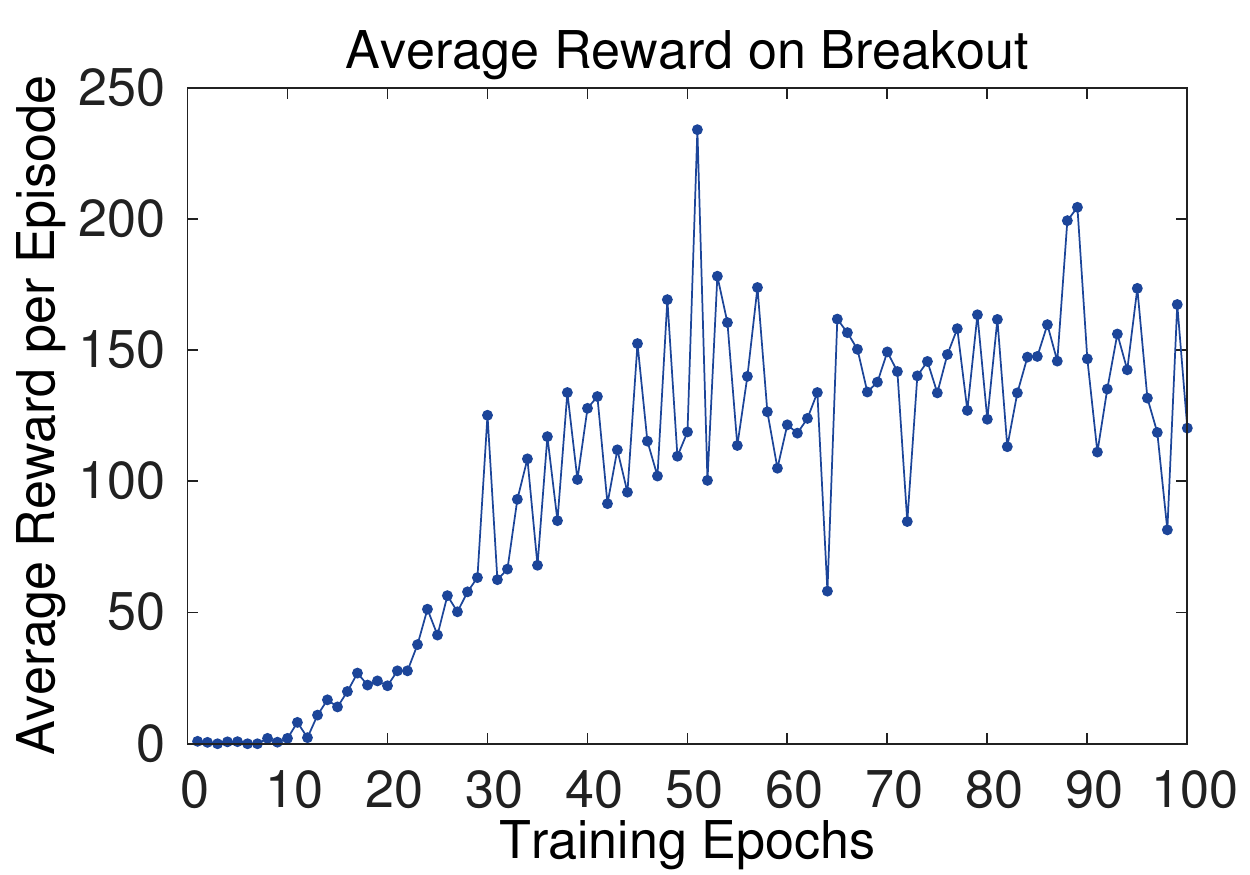}
\includegraphics[width=0.245\linewidth]{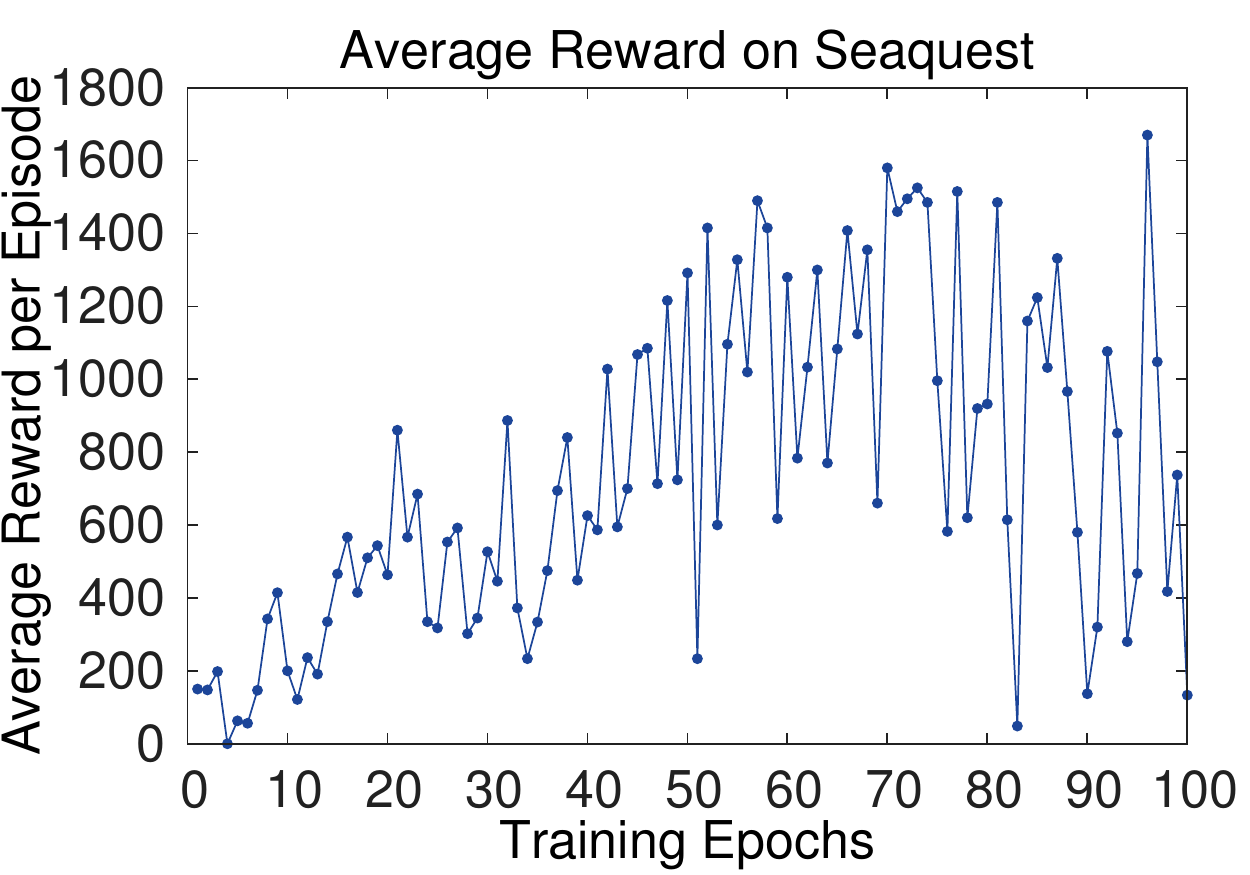}
\includegraphics[width=0.245\linewidth]{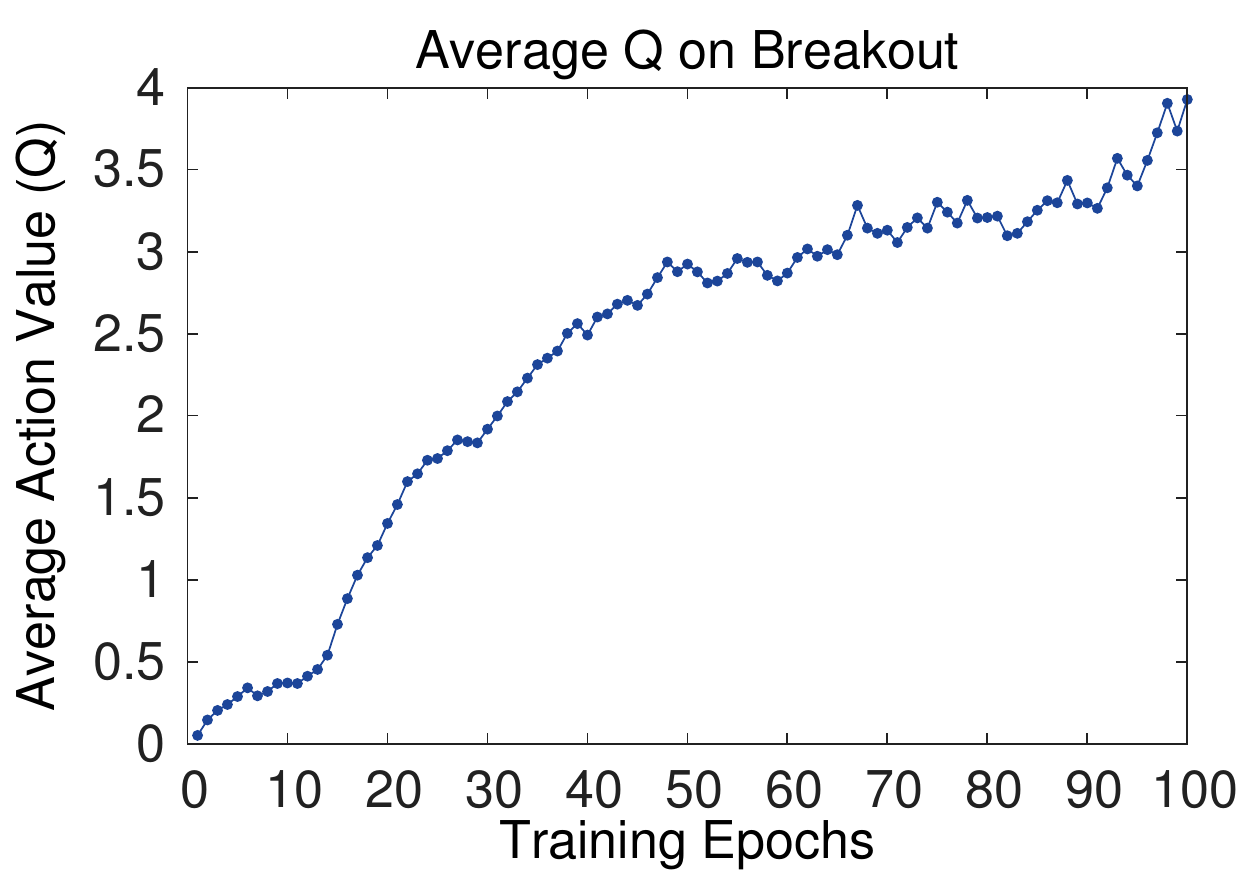}
\includegraphics[width=0.245\linewidth]{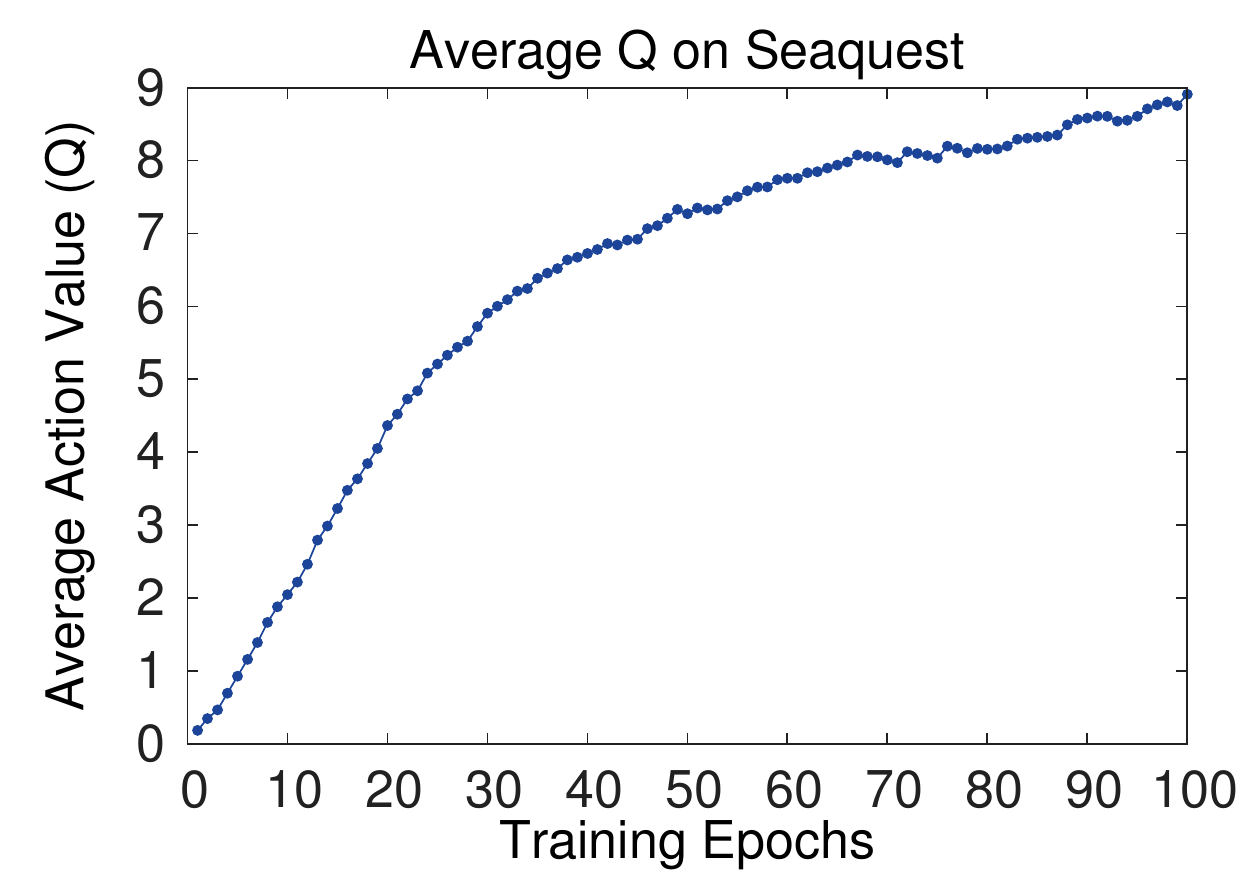}
\caption{\label{fig-V} The two plots on the left show average reward per episode on Breakout and Seaquest respectively during training.  The statistics were computed by running an $\epsilon$-greedy policy with $\epsilon=0.05$ for 10000 steps. The two plots on the right show the average maximum predicted action-value of a held out set of states on Breakout and Seaquest respectively.  One epoch corresponds to 50000 minibatch weight updates or roughly 30 minutes of training time.}
\end{figure}

In supervised learning, one can easily track the performance of a model during training by evaluating it on the training and validation sets. In reinforcement learning, however, accurately evaluating the progress of an agent during training can be challenging. 
Since our evaluation metric, as suggested by~\cite{bellemare-ale}, is the total reward the agent collects in an episode or game averaged over a number of games, we periodically compute it during training. The average total reward metric tends to be very noisy because small changes to the weights of a policy can lead to large changes in the distribution of states the policy visits .  The leftmost two plots in figure~\ref{fig-V} show how the average total reward evolves during training on the games Seaquest and Breakout.  Both averaged reward plots are indeed quite noisy, giving one the impression that the learning algorithm is not making steady progress.  Another, more stable, metric is the policy's estimated action-value function $Q$, which provides an estimate of how much discounted reward the agent can obtain by following its policy from any given state.  We collect a fixed set of states by running a random policy before training starts and track the average of the maximum\footnote{The maximum for each state is taken over the possible actions.} predicted $Q$ for these states.  The two rightmost plots in figure~\ref{fig-V} show that average predicted $Q$ increases much more smoothly than the average total reward obtained by the agent and plotting the same metrics on the other five games produces similarly smooth curves.
In addition to seeing relatively smooth improvement to predicted $Q$ during training we did not experience any divergence issues in any of our experiments.  This suggests that, despite lacking any theoretical convergence guarantees, our method is able to train large neural networks using a reinforcement learning signal and stochastic gradient descent in a stable manner.  


\subsection{Visualizing the Value Function}
\begin{figure}
\includegraphics[width=0.24\linewidth]{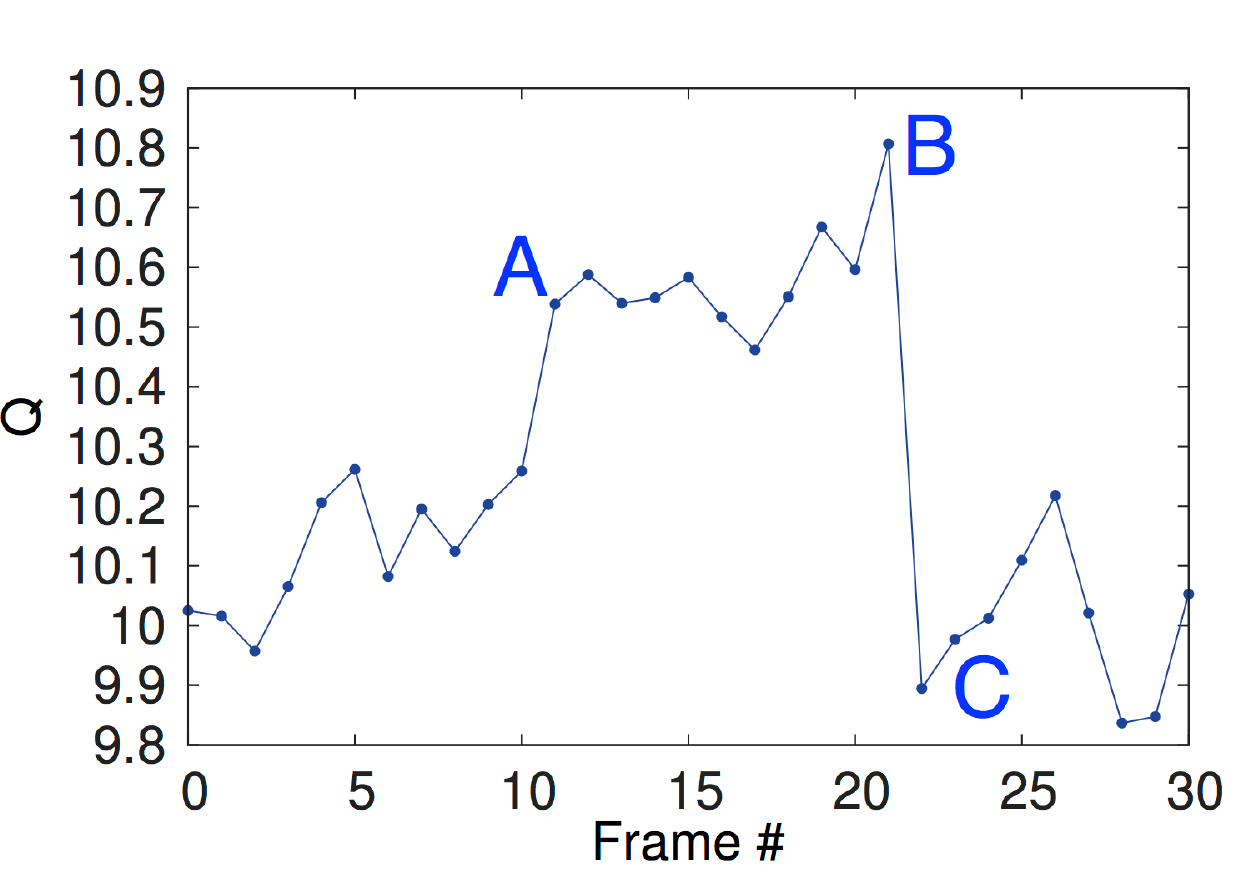}
\includegraphics[width=0.24\linewidth]{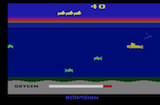}
\includegraphics[width=0.24\linewidth]{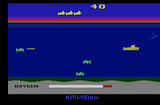}
\includegraphics[width=0.24\linewidth]{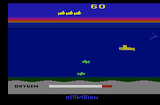}
\caption{\label{fig-reward} The leftmost plot shows the predicted value function for a 30 frame segment of the game Seaquest.  The three screenshots correspond to the frames labeled by A, B, and C respectively.}
\end{figure}

Figure~\ref{fig-reward} shows a visualization of the learned value function on the game Seaquest.  The figure shows that the predicted value jumps after an enemy appears on the left of the screen (point A). The agent then fires a torpedo at the enemy and the predicted value peaks as the torpedo is about to hit the enemy (point B). Finally, the value falls to roughly its original value after the enemy disappears (point C).  Figure~\ref{fig-reward} demonstrates that our method is able to learn how the value function evolves for a reasonably complex sequence of events. 

\subsection{Main Evaluation}

We compare our results with the best performing methods from the RL literature~\cite{bellemare-ale,bellemare-contingency}.  The method labeled \textbf{Sarsa} used the Sarsa algorithm to learn linear policies on several different feature sets hand-engineered for the Atari task and we report the score for the best performing feature set~\cite{bellemare-ale}.  \textbf{Contingency} used the same basic approach as \textbf{Sarsa} but augmented the feature sets with a learned representation of the parts of the screen that are under the agent's control \cite{bellemare-contingency}.  Note that both of these methods incorporate significant prior knowledge about the visual problem by using background subtraction and treating each of the 128 colors as a separate channel.  Since many of the Atari games use one distinct color for each type of object, treating each color as a separate channel can be similar to producing a separate binary map encoding the presence of each object type.  In contrast, our agents only receive the raw RGB screenshots as input and must \emph{learn} to detect objects on their own.   

In addition to the learned agents, we also report scores for an expert human game player and a policy that selects actions uniformly at random. The human performance is the median reward achieved after around two hours of playing each game. Note that our reported human scores are much higher than the ones in Bellemare et al.~\cite{bellemare-ale}.  For the learned methods, we follow the evaluation strategy used in Bellemare et al. \cite{bellemare-ale,bellemare:qtf} and report the average score obtained by running an $\epsilon$-greedy policy with $\epsilon=0.05$ for a fixed number of steps. The first five rows of table~\ref{table-results} show the per-game average scores on all games.  Our approach (labeled DQN) outperforms the other learning methods by a substantial margin on all seven games despite incorporating almost no prior knowledge about the inputs.   

\begin{table}
\begin{center}
\begin{tabular}{ |l | c | c | c | c | c | c | c |}
	\hline
	& {\small \textbf{B. Rider}} & {\small \textbf{Breakout}} & {\small \textbf{Enduro}} & {\small \textbf{Pong}} & {\small \textbf{Q*bert}} & {\small \textbf{Seaquest}} & {\small \textbf{S. Invaders}}\\
	\hline
	{\small \textbf{Random}}                                    & $354$ & $1.2$ & $0$ & $-20.4$ & $157$ & $110$ & $179$ \\
	\hline
	{\small \textbf{Sarsa}~\cite{bellemare-ale}}                & $996$ & $5.2$ & $129$ & $-19$ & $614$ & $665$ & $271$ \\
	\hline
	{\small \textbf{Contingency}~\cite{bellemare-contingency} } & $1743$ & $6$ & $159$ & $-17$ & $960$ & $723$ & $268$ \\
	\hline
	{\small \textbf{DQN} }                                      & $\textbf{4092}$ & $\textbf{168}$ & $\textbf{470}$ & $\textbf{20}$ & $\textbf{1952}$ & $\textbf{1705}$ & $\textbf{581}$ \\
	\hline
	{\small \textbf{Human} }                                    & $7456$ & $31$ & $368$ & $-3$ & $18900$ & $28010$ & $3690$ \\
	\hline
	\hline
	{\small \textbf{HNeat Best} \cite{hausknecht-neuro}}        & $3616$ & $52$ & $106$ & $19$ & $1800$ & $920$ & $\textbf{1720}$ \\
	\hline
	{\small \textbf{HNeat Pixel} \cite{hausknecht-neuro}}       & $1332$ & $4$ & $91$ & $-16$ & $1325$ & $800$ & $1145$ \\
	\hline
	{\small \textbf{DQN Best}}                                  & $\textbf{5184}$ & $\textbf{225}$ & $\textbf{661}$ & $\textbf{21}$ & $\textbf{4500}$ & $\textbf{1740}$ & $1075$ \\
	\hline
\end{tabular}
\caption{\label{table-results} The upper table compares average total reward for various learning methods by running an $\epsilon$-greedy policy with $\epsilon=0.05$ for a fixed number of steps. The lower table reports results of the single best performing episode for HNeat and DQN.  HNeat produces deterministic policies that always get the same score while DQN used an $\epsilon$-greedy policy with $\epsilon=0.05$.}
\end{center}
\end{table}

We also include a comparison to the evolutionary policy search approach from~\cite{hausknecht-neuro} in the last three rows of table~\ref{table-results}.   
We report two sets of results for this method. The \textbf{HNeat Best} score reflects the results obtained by using a hand-engineered object detector algorithm that outputs the locations and types of objects on the Atari screen. The \textbf{HNeat Pixel} score is obtained by using the special 8 color channel representation of the Atari emulator that represents an object label map at each channel. 
This method relies heavily on finding a deterministic sequence of states that represents a successful exploit. It is unlikely that strategies learnt in this way will generalize to random perturbations; therefore the algorithm was only evaluated on the highest scoring single episode. In contrast, our algorithm is evaluated on $\epsilon$-greedy control sequences, and must therefore generalize across a wide variety of possible situations.  Nevertheless, we show that on all the games, except Space Invaders, not only our max evaluation results (row $8$), but also our average results (row $4$) achieve better performance.

Finally, we show that our method achieves better performance than an expert human player on Breakout, Enduro and Pong and it achieves close to human performance on Beam Rider. The games Q*bert, Seaquest, Space Invaders, on which we are far from human performance, are more challenging because they require the network to find a strategy that extends over long time scales.

%% file: conclusion.tex
This paper introduced a new deep learning model for reinforcement learning, and demonstrated its ability to master difficult control policies for Atari 2600 computer games, using only raw pixels as input.
We also presented a variant of online Q-learning that combines stochastic minibatch updates with experience replay memory to ease the training of deep networks for RL.
Our approach gave state-of-the-art results in six of the seven games it was tested on, with no adjustment of the architecture or hyperparameters.

%% file: deepqnet.bbl
\begin{thebibliography}{10}

\bibitem{baird:residual}
Leemon Baird.
\newblock Residual algorithms: Reinforcement learning with function
  approximation.
\newblock In {\em Proceedings of the 12th International Conference on Machine
  Learning (ICML 1995)}, pages 30--37. Morgan Kaufmann, 1995.

\bibitem{bellemare2012sketch}
Marc Bellemare, Joel Veness, and Michael Bowling.
\newblock Sketch-based linear value function approximation.
\newblock In {\em Advances in Neural Information Processing Systems 25}, pages
  2222--2230, 2012.

\bibitem{bellemare-ale}
Marc~G Bellemare, Yavar Naddaf, Joel Veness, and Michael Bowling.
\newblock The arcade learning environment: An evaluation platform for general
  agents.
\newblock {\em Journal of Artificial Intelligence Research}, 47:253--279, 2013.

\bibitem{bellemare-contingency}
Marc~G Bellemare, Joel Veness, and Michael Bowling.
\newblock Investigating contingency awareness using atari 2600 games.
\newblock In {\em AAAI}, 2012.

\bibitem{bellemare:qtf}
Marc~G. Bellemare, Joel Veness, and Michael Bowling.
\newblock Bayesian learning of recursively factored environments.
\newblock In {\em Proceedings of the Thirtieth International Conference on
  Machine Learning (ICML 2013)}, pages 1211--1219, 2013.

\bibitem{dahl-speech}
George~E. Dahl, Dong Yu, Li~Deng, and Alex Acero.
\newblock Context-dependent pre-trained deep neural networks for
  large-vocabulary speech recognition.
\newblock {\em Audio, Speech, and Language Processing, IEEE Transactions on},
  20(1):30 --42, January 2012.

\bibitem{graves-speech}
Alex Graves, Abdel-rahman Mohamed, and Geoffrey~E. Hinton.
\newblock Speech recognition with deep recurrent neural networks.
\newblock In {\em Proc. ICASSP}, 2013.

\bibitem{hausknecht-neuro}
Matthew Hausknecht, Risto Miikkulainen, and Peter Stone.
\newblock A neuro-evolution approach to general atari game playing.
\newblock 2013.

\bibitem{heess:energy-based}
Nicolas Heess, David Silver, and Yee~Whye Teh.
\newblock Actor-critic reinforcement learning with energy-based policies.
\newblock In {\em European Workshop on Reinforcement Learning}, page~43, 2012.

\bibitem{jarrett-best}
Kevin Jarrett, Koray Kavukcuoglu, Marc’Aurelio Ranzato, and Yann LeCun.
\newblock What is the best multi-stage architecture for object recognition?
\newblock In {\em Proc. International Conference on Computer Vision and Pattern
  Recognition (CVPR 2009)}, pages 2146--2153. IEEE, 2009.

\bibitem{krizhevsky-imagenet}
Alex Krizhevsky, Ilya Sutskever, and Geoff Hinton.
\newblock Imagenet classification with deep convolutional neural networks.
\newblock In {\em Advances in Neural Information Processing Systems 25}, pages
  1106--1114, 2012.

\bibitem{lange:dfq}
Sascha Lange and Martin Riedmiller.
\newblock Deep auto-encoder neural networks in reinforcement learning.
\newblock In {\em Neural Networks (IJCNN), The 2010 International Joint
  Conference on}, pages 1--8. IEEE, 2010.

\bibitem{lin1993reinforcement}
Long-Ji Lin.
\newblock Reinforcement learning for robots using neural networks.
\newblock Technical report, DTIC Document, 1993.

\bibitem{maei:nonlinear}
Hamid Maei, Csaba Szepesvari, Shalabh Bhatnagar, Doina Precup, David Silver,
  and Rich Sutton.
\newblock {Convergent Temporal-Difference Learning with Arbitrary Smooth
  Function Approximation}.
\newblock In {\em Advances in Neural Information Processing Systems 22}, pages
  1204--1212, 2009.

\bibitem{maei:gq}
Hamid Maei, Csaba Szepesv{\'a}ri, Shalabh Bhatnagar, and Richard~S. Sutton.
\newblock Toward off-policy learning control with function approximation.
\newblock In {\em Proceedings of the 27th International Conference on Machine
  Learning (ICML 2010)}, pages 719--726, 2010.

\bibitem{mnih-thesis}
Volodymyr Mnih.
\newblock {\em Machine Learning for Aerial Image Labeling}.
\newblock PhD thesis, University of Toronto, 2013.

\bibitem{moore:prioritized}
Andrew Moore and Chris Atkeson.
\newblock Prioritized sweeping: Reinforcement learning with less data and less
  real time.
\newblock {\em Machine Learning}, 13:103--130, 1993.

\bibitem{nair-relu}
Vinod Nair and Geoffrey~E Hinton.
\newblock Rectified linear units improve restricted boltzmann machines.
\newblock In {\em Proceedings of the 27th International Conference on Machine
  Learning (ICML 2010)}, pages 807--814, 2010.

\bibitem{pollack:td-gammon}
Jordan~B. Pollack and Alan~D. Blair.
\newblock Why did td-gammon work.
\newblock In {\em Advances in Neural Information Processing Systems 9}, pages
  10--16, 1996.

\bibitem{riedmiller-nfq}
Martin Riedmiller.
\newblock Neural fitted q iteration--first experiences with a data efficient
  neural reinforcement learning method.
\newblock In {\em Machine Learning: ECML 2005}, pages 317--328. Springer, 2005.

\bibitem{sallans:rbm}
Brian Sallans and Geoffrey~E. Hinton.
\newblock Reinforcement learning with factored states and actions.
\newblock {\em Journal of Machine Learning Research}, 5:1063--1088, 2004.

\bibitem{sermanet-cvpr-2013}
Pierre Sermanet, Koray Kavukcuoglu, Soumith Chintala, and Yann LeCun.
\newblock Pedestrian detection with unsupervised multi-stage feature learning.
\newblock In {\em Proc. International Conference on Computer Vision and Pattern
  Recognition (CVPR 2013)}. IEEE, 2013.

\bibitem{sutton:book}
Richard Sutton and Andrew Barto.
\newblock {\em Reinforcement Learning: An Introduction}.
\newblock MIT Press, 1998.

\bibitem{tesauro-95-backgammon}
Gerald Tesauro.
\newblock Temporal difference learning and td-gammon.
\newblock {\em Communications of the ACM}, 38(3):58--68, 1995.

\bibitem{tsitsiklis:td-convergence}
John~N Tsitsiklis and Benjamin Van~Roy.
\newblock An analysis of temporal-difference learning with function
  approximation.
\newblock {\em Automatic Control, IEEE Transactions on}, 42(5):674--690, 1997.

\bibitem{watkins-qlearning}
Christopher~JCH Watkins and Peter Dayan.
\newblock Q-learning.
\newblock {\em Machine learning}, 8(3-4):279--292, 1992.

\end{thebibliography}
